# SkySim: A ROS2-based Simulation Environment for Natural Language Control of Drone Swarms using Large Language Models

**Aditya Shibu[1], Marah Saleh[2], Mohamed Al-Musleh[3], and Nidhal Abdulaziz[4]**

[1]School of Mathematical and Computer Sciences, [2,3]School of Textiles and Design (AIM BeyonD Research Lab), [4] School of Engineering and Physical Sciences, Heriot-Watt University Dubai Campus

E-mail: [1]as2397@hw.ac.uk, [2]Marah.Saleh@hw.ac.uk, [3]M.Al-Musleh@hw.ac.uk,
[4]Nidhal.Abdulaziz@hw.ac.uk.

**Summary:** Unmanned Aerial Vehicle (UAV) swarms offer versatile applications in logistics, agriculture, and surveillance, yet controlling them requires expert knowledge for safety and feasibility. Traditional static methods limit adaptability, while Large Language Models (LLMs) enable natural language control but generate unsafe trajectories due to lacking physical grounding. This paper introduces SkySim, a ROS2-based simulation framework in Gazebo that decouples LLM high-level planning from low-level safety enforcement. Using Gemini 3.5 Pro, SkySim translates user commands (e.g., "Form a circle") into spatial waypoints, informed by real-time drone states. An Artificial Potential Field (APF) safety filter applies minimal adjustments for collision avoidance, kinematic limits, and geo-fencing, ensuring feasible execution at 20 Hz.
Experiments with swarms of 3, 10, and 30 Crazyflie drones validate spatial reasoning accuracy (100% across tested geometric primitives), real-time collision prevention, and scalability. SkySim empowers non-experts to iteratively refine behaviors, bridging AI cognition with robotic safety for dynamic environments. Future work targets hardware integration.

## 1. Introduction

Unmanned Aerial Vehicle (UAV) swarms are powerful tools capable of performing tasks across diverse domains, including logistics, agriculture, and surveillance [1]. However, designing complex movements or controlling these systems remains a complex challenge, requiring significant expert knowledge to ensure the balance between safety and physical feasibility [2]. Historically, UAV systems have relied on static control methods such as pre-programmed task flows and rule-based libraries, which restrict them to executing predefined actions and limit their adaptability in dynamic, real-world environments [3]. The emergence of Large Language Models (LLMs) offers a potential solution by allowing users to translate high-level human intentions into actionable control commands via Natural Language [1][4]. Nevertheless, LLM-generated reference trajectories are inherently unsafe, as LLMs lack the real-world grounding necessary to understand physics or predict events like collisions, especially since even small errors in reasoning can lead to immediate failure in drone swarms [2][4].

To realize the benefits of natural language control while maintaining physical security, a decoupled mechanism is needed to enforce safety constraints. In this paper we introduce **SkySim**, a ROS2 based simulation framework that bridges this gap by decoupling the LLM's high-level planning from a low-level, real-time safety filter. This system leverages the LLM's reasoning power to control a swarm of drones, while an optimization-based safety filter operates beneath it, ensuring trajectories are feasible, collision free, and safe for real-world deployment by applying minimal adjustments where constraints are violated. This architecture enables non-experts to iteratively refine swarm behaviors using natural language, without the burden of accounting for safety or low-level feasibility constraints.

### 1.1. Related Work

The integration of Large Language Models (LLMs) into Unmanned Aerial Vehicle (UAV) systems has emerged as a transformative approach to bridging the gap between high-level human intentions and actionable control commands. Recent literature demonstrates that LLMs enable users to issue natural language instructions that can be converted into executable trajectories, moving past the limitations of traditional static frameworks [1-3]. However, the reliability challenges and lack of real-world grounding in LLM-generated outputs require the development of hybrid architectures to ensure safety in mission critical applications [4-6].

A leading paradigm in this domain is the SwarmGPT framework [7], which pioneered the integration of LLMs for synchronized drone swarm choreography. SwarmGPT addresses fundamental safety challenges by augmenting the LLM with an optimization-based safety filter that minimally corrects generated trajectories to ensure they remain collision-free. However, this architecture is largely optimized for temporally coordinated choreography, focusing on the smoothness and aesthetic quality of the swarm's motion over time. It is less focused on zero-shot spatial reasoning, where the primary objective is to rapidly translate abstract natural language commands into valid, arbitrary static geometries (e.g., 'form a Christmas Tree') without prior training on those specific shapes. Consequently, alternative strategies focusing on real-time reachable set generation and spatial optimization have emerged as crucial complements, highlighting the need for frameworks that can ensure safety during rapid, goal-oriented reconfiguration [8], [9].

Furthermore, the effectiveness of the safety filter relies heavily on the choice of collision avoidance algorithm [10]. While recent research highlights the efficacy of techniques such as Convolutional Neural Networks (CNNs) and Deep Reinforcement Learning (DRL) [11-14], these approaches often add significant computational overhead and training complexity. In an architecture where the high-level planner (LLM) is already subject to variable inference latency, adding a computationally intensive learning-based safety layer increases the risk of control loop delays, potentially leading to collisions during rapid and clustered maneuvers. In contrast, deterministic reactive models like Artificial Potential Fields (APF) offer a lightweight, distinct alternative [11]. By leveraging real-time state information, readily available via ground truth or motion capture systems, APF provides immediate, reflexive collision avoidance with negligible computational cost. This validates the decoupled "sense-plan-act" architecture that SkySim aims to achieve, where the LLM handles the high-level cognitive reasoning while a fast, deterministic APF layer enforces physical safety constraints in real-time.

## 2. System Architecture and Methodology

Our methodology is established on a modular, decoupled architecture implemented within the Robot Operating System (ROS2 Jazzy). To ensure system stability under variable network conditions, the framework strictly separates high-latency cognitive planning from low-latency real-time execution.

Fig. 1 illustrates this multi-layered architecture, detailing the asynchronous data flow between the Planning Layer (LLM-driven goal generation) and the Control Layer (20Hz safety enforcement).

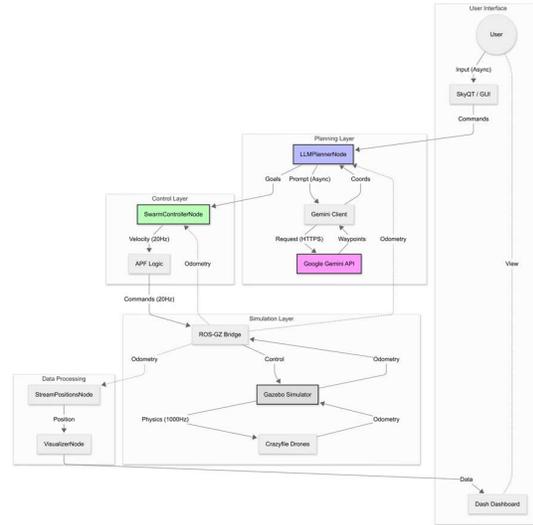

**Fig. 1.** SkySim system architecture.

### 2.1. Command Input and State Sensing

The process is initiated via a natural language command (e.g., "Form a circle around the center at a 2m height"). Upon receiving the input, the SkySim interface initiates the "sense" phase by subscribing to the current spatial coordinates of all agents via the ROS2 (/odom) topic.

The (/odom) topic is bridged directly from the Gazebo Simulation to the control node. This is done to emulate the high-precision, real-world motion capture systems (e.g., OptiTrack). This provides a noiseless, instantaneous snapshot of the current swarm topology.

### 2.2. State-Aware LLM-based Planning

Contrary to standard LLM control pipelines, SkySim does not attempt to map text directly to discrete control primitives (e.g., throttle, pitch). Instead, it acts as a high-level spatial planner via the Gemini 3.5 Pro LLM (google.generativeai API).

To ensure deterministic and physically feasible outputs, the LLM is initialized with strict system instruction defining the swarm size *(N),* the coordinate system *(X=forward, Y=left, Z=up),* and the ground-collision constraints *(Z ≥ 0.5 m).*

When a user issues a command, the framework constructs a dynamic state-aware prompt by concatenating the system instructions, the instantaneous 3D positional telemetry of all the drones, and the natural language input.

**Example Prompt Structure (N = 3):**
**System Instruction:**
"You are a drone swarm controller for 3 drones. Generate target [x, y, z] coordinates to fulfill the command. Output only a valid python list of 3 lists. Keep *Z ≥ 0.5".*

**Current Context:**
Current Drone Positions:
[[0.1, 0.0, 1.0], [0.0, 1.5, 1.1], [1.2, 1.1, 0.9]].
**User Command:**
"Form a triangle around the center."

The LLM is constrained to output a serialized configuration matrix, formatted as a raw Python list as shown in Eq 1:

$$P_{goal} = [[x_1, y_1, z_1], ..., [x_N, y_N, z_N]] \quad (1)$$

This string is deterministically parsed via Abstract Syntax Tree for validation. No additional formatting, markdown, or text explanations are permitted. In the event of an API timeout or malformed output schema, the system holds the position of the previous command to prevent undefined system states.

## 2.2. The Safety Filter and Safe Execution

The "act" phase is managed by the Safety Filter, which bridges the gap between static LLM waypoints and real-time dynamic feasibility. We model the multi-agent system using a single-integrator kinematic model, where the control input for agent $i$ is its velocity vector, $u_i = \dot{p}_i$.

To enforce collision avoidance in real-time without the computational overhead of non-linear solvers (e.g., MPC), SkySim implements an Artificial Potential Field controller. The control loop operates at 20 Hz, continuously calculating the required velocity as a gradient of a composite potential function:

$$v_{total,i} = v_{att,i} + \sum_{j \neq i} v_{rep,ij} \quad (2)$$

**1. Attractive Velocity (Goal Seeking):** A proportional controller puts agent $i$ toward the LLM-assigned waypoint $P_{goal,i}$ with a gain of $K_p = 1.0$:

$$v_{att,i} = K_p \cdot (P_{goal,i} - P_i) \quad (3)$$

**2. Repulsive Velocity (Collision Avoidance):** A hard safety constraint is enforced to prevent inter-agent collisions. If the Euclidean distance $d_{ij} = \|P_i - P_j\|$ between agent $i$ and neighbour $j$ falls below the safety radius $r_{min} = 0.8$ m (can be configured), a linear repulsive vector is activated:

$$v_{(rep,ij)} = \begin{cases} K_{rep} \cdot (r_{min} - d_{ij}) \cdot \hat{n}_{ij} & if\ d_{ij} < r_{min} \\ 0 & otherwise \end{cases} \quad (4)$$

Where $\hat{n}_{ij} = \frac{P_i - P_j}{d_{ij}}$ is the unit vector pointing away from the neighbour, and $K_{rep} = 2.0$ is the repulsion gain.

**3. Kinematic and Boundary Constraints:** To ensure absolute system safety, commands are sanitized via a two-stage constraint protocol before execution. First, the LLM Planner Node enforces a strict virtual geo-fence. Any LLM-generated waypoint that falls outside the defined spatial boundaries ($X, Y \in [10.0, 10.0]\ m, Z \in [0.2, 5.0]\ m$) is immediately rejected, triggering a position-hold fallback to prevent the swarm from exiting the simulated range of a motion capture system.

Secondly, for waypoints within a valid workspace, the kinematic limits of the nano-quadrotors are enforced. The final resultant velocity command is saturated at $v_{max} = 0.5$ m/s before being published to the ROS2 (/cmd_vel) topic:

$$v_{cmd,i} = \min(|v_{total,i}|, v_{max}) \cdot \frac{v_{total,i}}{|v_{total,i}|} \quad (5)$$

This multi-tiered architecture guarantees that the swarm remains bounded within a safe operational volume and respects the physical motor limits at all times.

## 3. Experimental Evaluation and Results

To validate the safety, scalability, and cognitive capabilities of the SkySim framework, we conducted a series of simulation experiments using Gazebo Harmonic Environment and ROS2 Jazzy. The swarm consists of simulated Crazyflie 2.1 Brushless. The positions of the drones are retrieved directly from the ground truth provided by Gazebo under the ROS2 (/odom) topic.

The evaluation focuses on three primary metrics: LLM spatial reasoning accuracy, real-time collision avoidance, and system scalability. Swarm sizes of (N = 3, 10, and 30) agents were tested to evaluate the framework under increased density.

### 3.1. Test Scenarios and Swarm Scalability

To evaluate the framework, we designed three distinct experimental scenarios to test LLM spatial reasoning, real-time safety, and boundary enforcement. The system was dynamically scaled across swarm sizes of ($N$ = 3, 10, and 30) agents to evaluate the computational limits of the APF controller, the LLM context window, and the scalability aspect of the framework.

**Scenario A: Geometric Formation Control (Spatial Reasoning)**
The swarm was commanded to perform high-level topological tasks, including: "Form a circle with a radius of 2 meters.", "Form a sphere of radius 2 meters.", "Form a 5x6 grid at a height of 2 meters.", "Form a 3D cube.", and "Form a christmas tree.".

Gemini 3.5 Pro Preview successfully translated these abstract commands into valid *N*-dimensional spatial coordinates. Across all swarm sizes (*N* = 3 to 30), the LLM achieved a 100% success rate on the defined geometric primitives, accurately calculating target coordinates for every test conducted. The system demonstrated robust scalability, with the LLM generating actionable plans within ≈ 50 seconds (for N=30) and the APF controller successfully converging the agents to their target formations without collision.

Fig. 2. Drone formations generated via LLM prompts: (left) N = 3 triangle, (center) N = 10 circle with a radius 3 m, (right) N = 30 5×6 grid. Prompts:

- ***N* = 3:** "Form a triangle at height 2m centered at zero."
- ***N* = 10:** "Form a circle of radius 3m at height 2m centered at zero."
- ***N* = 30:** *"Form a 5x6 grid pattern at height 2m."*

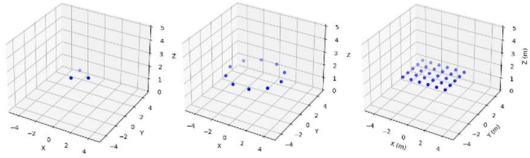

**Fig. 2.** The final positions of the drones after reaching the waypoints set by the LLM

**Scenario B: Static and Dynamic Collision Avoidance (APF Stress-Tests)**

To evaluate the operational limits of the decoupled APF controller, we tested the medium swarm size (*N* = 10) under two distinct collision-avoidance tests.

**The Static Hazard (Unsafe Waypoint Test):** An automated stress-test was triggered, commanding all of the swarm to converge on to a single point. Fig. 3. Shows the inter-agent separation distances. As illustrated, there is only a small moment where the Euclidean distance marginally drops below the 0.8-meter threshold. This brief moment of the safety boundary is the exact mechanism that triggers the APF, the resultant repulsive vector instantly dominated the attractive goal-seeking force. This shows that the APF successfully detected when the drones were converging and the repulsion forces dominated the attractive forces.

**The Dynamic Hazard (Positional Swap Test):** The swarm was given a natural language command "All drones swap positions with the drone directly opposite to you." This command forces all agent trajectories to dynamically intersect simultaneously at the center of the workspace. Without the low-level safety filter, this would result in a catastrophic multi-agent collision. However, the repulsive vector fields ($r_{min}$ = 0.8) dynamically routed the agents around each other in real-time, allowing the swarm to resolve the conflict and reach their target waypoints safely.

**3.2. Safety and Feasibility Metrics**

To strictly quantify the performance of the safety filter, we monitored two key metrics: Minimum Inter-Agent Distance ($d_{min}$) for collision verification and Swarm Velocity Profile across all unique agent pairs throughout the simulation runtime. We analyzed this metric specifically during the formation of high-density structures: the *N* = 10 "Christmas Tree", N = 30 "Christmas Tree" (Complex 3D topology), and the N = 30 "Star" (high-density planar topology).

Fig. 3 presents the resulting formation snapshots alongside their distance profiles. It is important to note that the red dashed line at 0.8m represents the APF activation threshold, not the physical collision boundary. As observed, $d_{min}$ naturally dips below this conservative buffer to stabilize the dense formations, yet it remains strictly above the physical collision limit ($r_{drone}$ ≈ 55mm [15]) at all times, validating real-time safety."

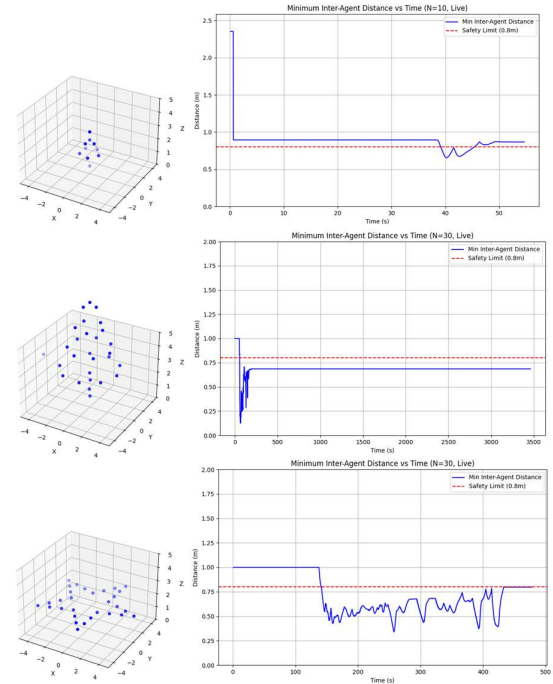

**Fig. 3.** (Left) 3D swarm snapshots. (Right) Minimum inter-agent distance over time. The red line (0.8m) indicates the APF activation buffer, no physical collisions occurred.

To ensure the generated trajectories were smooth and flyable, we analyzed the velocity profile of the N=10 swarm during reconfiguration, as shown in Fig. 4. The profile demonstrates a smooth acceleration phase followed by a natural decay, with no high-frequency oscillations. the peak velocity (≈ 0.42 m/s) remains strictly within the defined kinematic limit ($v_{max}$ = 0.5 m/s), validating that the decoupled architecture produces physically feasible control commands.

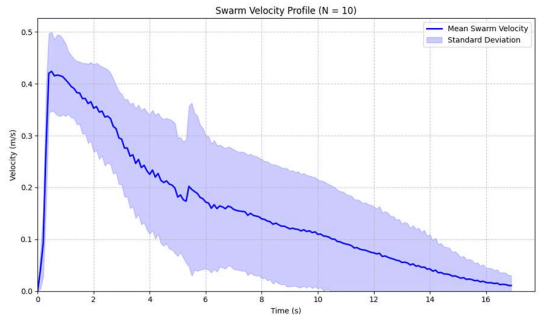

**Fig. 4.** Velocity Profile of drones in N = 10 configuration

### 3.3. Latency Analysis

To evaluate the computational viability of the framework for large-scale operations, we measured the End-to-End planning latency, defined as the total time elapsed between issuing a natural language command and receiving the parsed coordinate list from the LLM.

Fig. 5. illustrates the latency distribution across varying swarm sizes ($N$ = 3, 4, 10, 30)

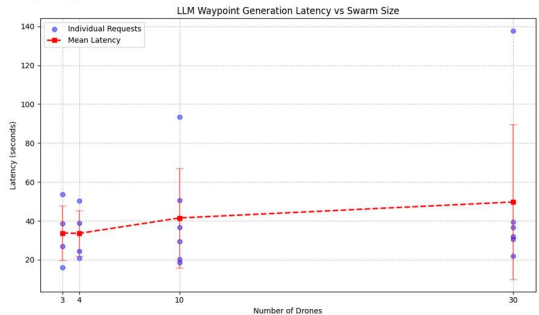

**Fig. 5.** End-to-end planning latency vs. Swarm Size.

The results highlight two critical performance characteristics:

**Sub-Linear Scalability:** The system demonstrates scalability with respect to swarm density. While the swarm size increased by a factor of 10 ($N$ = 3 to $N$ = 10), the mean planning latency only increased by approximately 45% (34s to 50s). This indicates that the computational cost of processing the drone state data increases efficiently, proving that the system can handle larger swarms without a drastic increase in processing time.

**Decoupled Safety:** Significant variance is observed in the planning time (e.g., outliers at $N$ = 30), which in a characteristic of cloud-based probabilistic models influenced by API server load and network jitter. However, this Planning Latency (seconds) is completely decoupled from the Control Loop Latency (milliseconds). As detailed in Fig. 1., the local safety filter operates continuously at 20Hz, ensuring that these variable planning delays never compromise the physical safety or stability of the swarm during the planning phase.

## 4. Discussion and Limitations

The experimental results validate SkySim's ability to translate natural language into safe, feasible swarm behaviors. Specifically, the 100% success rate on tested geometric primitives demonstrates the LLMs's spatial reasoning, effectively outperforming static control libraries in terms of adaptability and ease of use.

**Implications:** The integration of the decoupled safety filter proved essential. While the LLM generated valid goal states, the brief dips in inter-agent distances highlight the APF's reactive nature. This mechanism effectively prevented collision across all $N$ = 3, $N$ = 10, and $N$ = 30 trials, although at the cost of increased convergence time during high-density maneuvers. By decoupling high-level planning from low-level safety, SkySim mitigates risks in safety-critical domains, potentially allowing UAV applications in agriculture or disaster response where non-experts require complex, on-demand swarm behaviors without extensive programming knowledge.

**Limitations:** Despite these successes, several limitations in the current architecture must be addressed. The system's reliance on cloud-based inference (Google Gemini API). As shown in the latency analysis (Fig. 5), this introduces significant variance (outliers > 100s) driven by server load. While the local safety guarantees physical safety during these delays, the system is currently unsuitable for time-critical missions where millisecond-level reaction to new commands is required.

SkySim also relies on noiseless ground-truth state estimation from Gazebo. Real-world deployment will introduce sensor noise (e.g., from Optitrack [16] or GPS), communication packet loss, and environmental disturbances (e.g., weather conditions), which may degrade the precision of the APF repulsion vectors.

While the system scales sub-linearly, the LLM's context window limits the maximum swarm size. For $N$ > 30, the prompt length required to encode every agent's position may exceed token limits, our validation was capped at $N$ = 30 due to the local hardware resources required to maintain real-time fidelity for the physics engine and ROS2 node communication across all agents.

The deterministic APF controller, while fast is susceptible to local minima. In highly cluttered environments or complex obstacle fields, agents may become trapped in state equilibrium points before reaching their goal, a known limitation of reactive potential fields compared to predictive solvers.

Finally, the system is optimized for zero-shot static formations. It does not natively support continuous, time-bound trajectories (e.g., a "spinning sphere"), limiting its current scope of waypoint-based reconfiguration.

## 5. Conclusion and Future Works

This paper presented SkySim, a ROS2-based simulation environment that bridges the gap between natural language and LLM based cognition. By isolating the probabilistic LLM reasoning from a deterministic 20Hz safety filter, we demonstrated that non-experts can command complex swarm formations without compromising physical safety.

Future work will aim to focus on three key areas to address these identified limitations

**1. Hardware Integration:** We aim to transfer the stack to physical Crazyflie 2.1 Brushless drones [15] using OptiTrack motion capture system [16] to quantify the impact of real-world sensor noise.

**2. Advanced Control Solvers:** To overcome the local minima limitations of the APF, we will explore integrating Model Predictive Control (MPC) as a proactive safety layer that can handle dynamic constraints more effectively than a simple APF.

**3. Multi-Modal Inputs:** Future iterations will incorporate multi-modal capabilities, allowing the swarm to react to visual inputs or dynamic environmental changes, further extending the "sense-plan-act" loop beyond static coordinates.

**4. Edge Deployment and Optimization:** To eliminate the latency bottlenecks inherent to cloud-based inference, future iterations will transition from general-purpose cloud LLMs to Small Language Models (SLMs) deployed directly on local hardware. We plan to leverage efficient fine-tuning frameworks, such as Unsloth [17], to optimize these models specifically for swarm control vocabularies. This shift aims to reduce the end-to-end planning to sub-second levels, enabling the system in combination with MPC to support dynamic, high-speed maneuvers that are currently impossible with API-based delays.

## References

[1]. G. Aikins, M. Dao, J. Moukpe, T. Eskridge, and K. Nguyen, "LEVIOSA: Natural Language-Based Uncrewed Aerial Vehicle Trajectory Generation," Electronics, vol. 13, no. 22, p. 4508, 2024.

[2]. Y. Ping et al., "Multimodal Large Language Models-Enabled UAV Swarm: Towards Efficient and Intelligent Autonomous Aerial Systems," arXiv preprint arXiv:2506.12710, 2025.

[3]. B. Han, Y. Chen, J. Li, J. Li, and J. Su, "SwarmChain: Collaborative LLM Inference for UAV Swarm Control," IEEE Internet of Things Magazine, vol. 8, no. 5, pp. 64-71, Sept. 2025.

[4]. L. Yuan, C. Deng, D. Han, I. Hwang, S. Brunswicker, and C. Brinton, "Next-Generation LLM for UAV: From Natural Language to Autonomous Flight," arXiv preprint arXiv:2510.21739, 2025.

[5]. S. Sarkar et al., "Review of LLM based Control," 2025, doi: 10.13140/RG.2.2.28595.54567.

[6]. H. Kurunathan, H. Huang, K. Li, W. Ni, and E. Hossain, "Machine Learning-Aided Operations and Communications of Unmanned Aerial Vehicles: A Contemporary Survey," IEEE Communications Surveys & Tutorials, vol. 26, no. 1, pp. 496-533, 2024.

[7]. M. Schuck et al., "SwarmGPT: Combining Large Language Models with Safe Motion Planning for Drone Swarm Choreography," arXiv preprint arXiv:2412.08428, 2025.

[8]. P. Xie, S. Diaconescu, F. Stoican, and A. Alanwar, "Roundabout Constrained Convex Generators: A Unified Framework for Multiply-Connected Reachable Sets," arXiv preprint arXiv:2511.07330, 2025.

[9]. Z. Wang and H. Ding, "Opposition-Based Learning Equilibrium Optimizer with Application in Mobile Robot Path Planning," International Journal of Robotics and Automation Technology, vol. 10, pp. 64-74, 2025.

[10]. P. Mishra, B. Boopal, and N. Mishra, "Real-Time 3D Routing Optimization for Unmanned Aerial Vehicle using Machine Learning," ICST Transactions on Scalable Information Systems, 2024.

[11]. J. Yasin, S. Mohamed, M. Haghbayan, J. Heikkonen, H. Tenhunen, and J. Plosila, "Unmanned Aerial Vehicles (UAVs): Collision Avoidance Systems and Approaches," IEEE Access, vol. 8, pp. 105139-105155, 2020.

[12]. X. Dai, Y. Mao, T. Huang, N. Qin, D. Huang, and Y. Li, "Automatic obstacle avoidance of quadrotor UAV via CNN-based learning," Neurocomputing, vol. 402, pp. 346-358, 2020.

[13]. R. Bălașa, M. C. Bîlu, and C. Iordache, "A Proximal Policy Optimization Reinforcement Learning Approach to Unmanned Aerial Vehicles Attitude Control," Land Forces Academy Review, vol. 27, no. 4, pp. 400-410, 2022.

[14]. J. Hu, L. Wang, T. Hu, C. Guo, and Y. Wang, "Autonomous Maneuver Decision Making of Dual-UAV Cooperative Air Combat Based on Deep Reinforcement Learning," Electronics, vol. 11, no. 3, p. 467, 2022.

[15]. Bitcraze AB, "Crazyflie 2.1 Brushless Datasheet," Malmö, Skåne County, Sweden [Online]. Available: https://www.bitcraze.io/documentation/hardware/crazyflie_2_1_brushless/crazyflie_2_1_brushless-datasheet.pdf. [Accessed: Jan. 24, 2026].

[16]. NaturalPoint, Inc., "PrimeX 22 Technical Specifications," Corvallis, OR, USA. [Online]. Available: https://optitrack.com/cameras/primex-22. [Accessed: Jan. 24, 2026].

[17]. Unsloth AI, "Unsloth Documentation," [Online]. Available: https://unsloth.ai/docs. [Accessed: Jan. 24, 2026].